%% file: main.tex
\documentclass[10pt,twocolumn,letterpaper]{article}

\usepackage[pagenumbers]{cvpr} % To force page numbers, e.g. for an arXiv version

\input{preamble}
\definecolor{cvprblue}{rgb}{0.21,0.49,0.74}
\usepackage[pagebackref,breaklinks,colorlinks,allcolors=cvprblue]{hyperref}

\usepackage{graphicx}
\usepackage{amsmath, amsthm}
\usepackage{amssymb}
\usepackage{booktabs}
\usepackage{enumitem}
\usepackage{mathtools}

\usepackage{multirow}
\usepackage{xcolor}
\usepackage{colortbl}
\usepackage{makecell}
\usepackage{soul}
\usepackage{pifont}

\newcounter{todocounter}

\definecolor{table_head}{RGB}{231, 240, 241}
\definecolor{table_col}{RGB}{242, 242, 242}
\definecolor{table_our}{RGB}{255, 240, 193}

\usepackage{algorithm}
\usepackage{algorithmic}
\def\paperID{} % *** Enter the Paper ID here
\def\confName{CVPR}
\def\confYear{2026}

\title{Meta-FC: Meta-Learning with Feature Consistency for Robust and Generalizable Watermarking}

\makeatletter
\newcommand{\printfnsymbol}[1]{%
  \textsuperscript{\@fnsymbol{#1}}%
}
\makeatother

\author{Yuheng Li\textsuperscript{\rm 1}, 
Weitong Chen\textsuperscript{\rm 1}\thanks{Corresponding author.}, 
Chengcheng Zhu\textsuperscript{\rm 2}, 
Jiale Zhang\textsuperscript{\rm 1},
Chunpeng Ge\textsuperscript{\rm 3},
Di Wu\textsuperscript{\rm 4},
Guodong Long\textsuperscript{\rm 5}\\
\textsuperscript{\rm 1} Yangzhou University \quad
\textsuperscript{\rm 2} Nanjing University \quad
\textsuperscript{\rm 3} Shandong University \\
\textsuperscript{\rm 4} La Trobe University \quad
\textsuperscript{\rm 5} University of Technology Sydney
\\
\tt\small MX120240572@stu.yzu.edu.cn \quad
\tt\small \{wtchen,jialezhang\}@yzu.edu.cn \quad
\tt\small chengchengzhu@smail.nju.edu.cn \\
\tt\small chunpeng\_ge@sdu.edu.cn \quad
\tt\small d.wu@latrobe.edu.au \quad
\tt\small guodong.long@uts.edu.au 
}

\begin{document}
\maketitle
\input{sec/0_abstract}    
\input{sec/1_intro}

{
    \small
    \bibliographystyle{ieeenat_fullname}
    \bibliography{main}
}

\end{document}

%% file: sec/0_abstract.tex
\begin{abstract}
Deep learning-based watermarking has made remarkable progress in recent years. 
To achieve robustness against various distortions, current methods commonly adopt a training strategy where a \underline{\textbf{s}}ingle \underline{\textbf{r}}andom \underline{\textbf{d}}istortion (SRD) is chosen as the noise layer in each training batch. 
However, the SRD strategy treats distortions independently within each batch, neglecting the inherent relationships among different types of distortions and causing optimization conflicts across batches.
As a result, the robustness and generalizability of the watermarking model are limited. 
To address this issue, we propose a novel training strategy that enhances robustness and generalization via \underline{\textbf{meta}}-learning with \underline{\textbf{f}}eature \underline{\textbf{c}}onsistency (Meta-FC). 
Specifically, we randomly sample multiple distortions from the noise pool to construct a meta-training task, while holding out one distortion as a simulated ``unknown'' distortion for the meta-testing phase.
Through meta-learning, the model is encouraged to identify and utilize neurons that exhibit stable activations across different types of distortions, mitigating the optimization conflicts caused by the random sampling of diverse distortions in each batch.
To further promote the transformation of stable activations into distortion-invariant representations, we introduce a feature consistency loss that constrains the decoded features of the same image subjected to different distortions to remain consistent.
Extensive experiments demonstrate that, compared to the SRD training strategy, Meta-FC improves the robustness and generalization of various watermarking models by an average of 1.59\%, 4.71\%, and 2.38\% under high-intensity, combined, and unknown distortions.
\end{abstract}

%% file: sec/1_intro.tex
\section{Introduction}
\label{sec:intro}

Robust digital watermarking is a crucial technology for copyright protection \cite{hsu1999hidden, andalibi2015digital, chang2017features} and has been extensively researched. With the rapid progress of deep learning, a growing number of watermarking approaches based on deep neural networks (DNNs) have been developed. These methods \cite{zhu2018hidden, ahmadi2020redmark, fang2022pimog, Fang2023denol, Wang2024must} typically adopt an end-to-end framework that consists of an \underline{\textbf{e}}ncoder, a \underline{\textbf{n}}oise layer, and a \underline{\textbf{d}}ecoder (END). The encoder embeds the watermark message into the cover image, while the decoder aims to extract and reconstruct the watermark message from the distorted images. The noise layer simulates various distortions during training to enhance the robustness of the model.

\begin{figure}[!t]
\centering
\includegraphics[width=0.95\columnwidth]{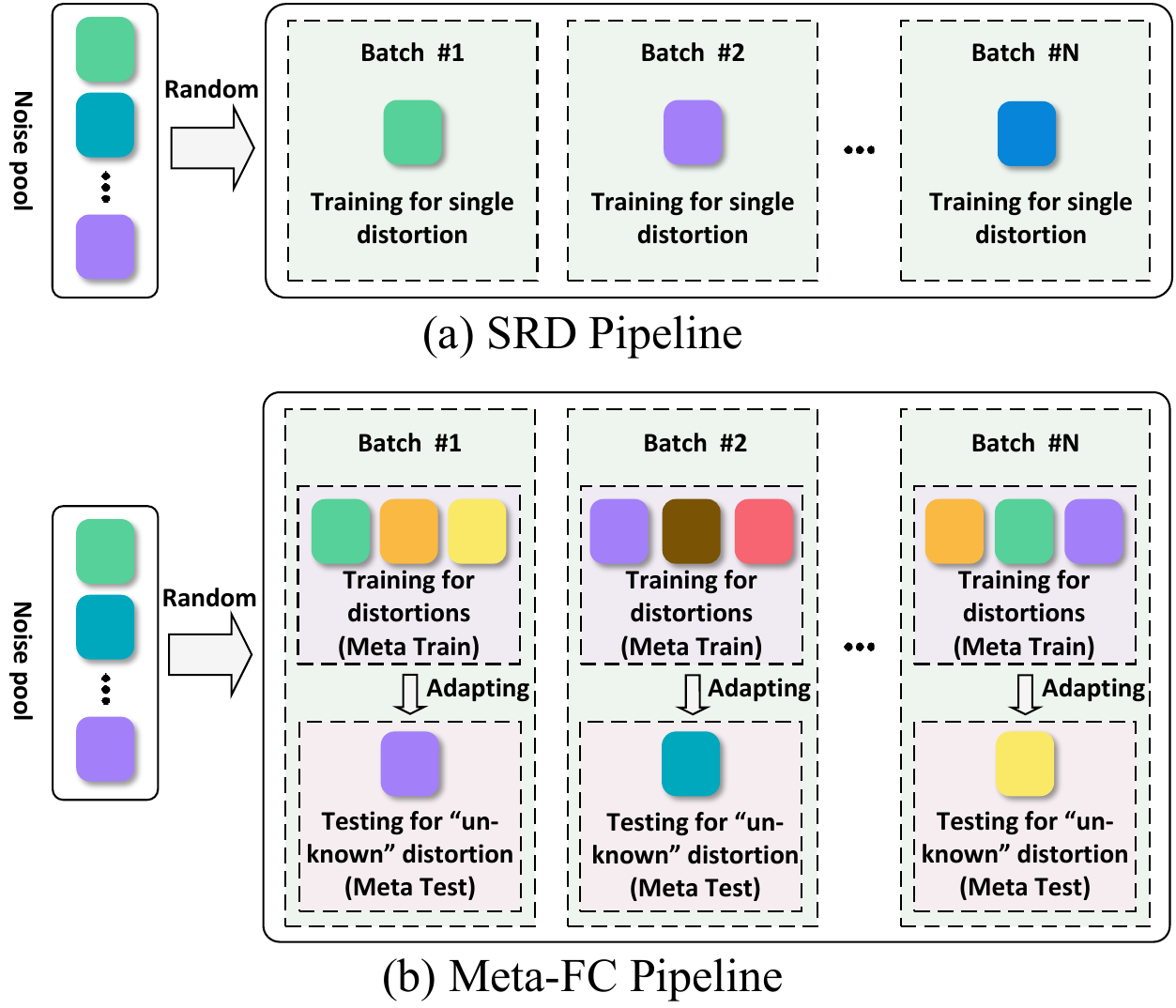}
\caption{The difference of the training process between SRD and Meta-FC. (a) The SRD pipeline. (b) The Meta-FC pipeline. In the meta-training phase, the model learns to be robust against known distortions. In the meta-testing phase, the model is evaluated under a simulated ``unknown” distortion. Note that no truly unknown distortions are involved during the entire training process.}
\label{fig:figure1}
\end{figure}

To advance the capabilities of deep learning-based watermarking, current research has largely centered on refining network architectures \cite{Fang2023De-END, ma2024geometric, kou2025iwrn} or developing sophisticated optimization objectives \cite{fernandez2022watermarking, Fang2022encoded, fang2024dero, liu2025watermarking, wu2025sim}. Although these approaches have achieved notable improvements, their impact on model robustness and generalization against diverse distortions remains constrained. We identify a critical, yet often overlooked, factor contributing to this limitation: the prevalent reliance on the \underline{\textbf{s}}ingle \underline{\textbf{r}}andom \underline{\textbf{d}}istortion (SRD) training strategy \cite{zhu2018hidden, liu2019novel, jia2021mbrs, Guo2023practical, fang2023flow}. Under the SRD paradigm, a single distortion is randomly selected from a predefined noise pool (e.g., JPEG, Crop) for each training batch, as illustrated in Figure \ref{fig:figure1} (a). However, this batch-by-batch, single-distortion approach inherently isolates the learning process for each distortion type. We argue that this isolation leads to two primary issues: 1) overfitting to distortion-specific features rather than learning true distortion-invariant representations, and 2) optimization instability, such as gradient conflicts across different distortions, which prevents the model from capturing these invariances. Consequently, the performance of watermarking models becomes demonstrably limited in three critical scenarios: \textit{high-intensity distortions}, \textit{combined distortions}, and \textit{unknown distortions} (i.e., those not encountered during training).

To address these limitations, we propose a novel training strategy for watermarking, called Meta-FC, which integrates \underline{\textbf{meta}}-learning with a \underline{\textbf{f}}eature \underline{\textbf{c}}onsistency loss.  
The core idea of Meta-FC is to improve the generalization of watermarking models by simulating training on known distortions and testing on ``unknown'' distortions within each training batch. This enables the model to discover distortion-invariant representations that are robust across diverse distortions.
Specifically, as illustrated in Figure~\ref{fig:figure1}(b), for each batch, we randomly sample a subset of distortions from a predefined noise pool as the meta-training distortions, while a held-out distortion is designated as the meta-testing distortion. During the meta-training phase, the model is optimized across multiple sampled distortions to obtain temporary encoder and decoder parameters, along with the corresponding meta-training loss. 
The temporary parameters are then evaluated on a meta-testing distortion, which is excluded from meta-training and simulated as an ``unknown'' distortion, to compute the meta-testing loss.
By jointly minimizing both losses, the model is encouraged to learn stable and adaptable parameters that remain effective under diverse and unknown distortion conditions. 
To further transform such stable and adaptable parameters into distortion-invariant representations, we introduce a feature consistency loss, which aligns the last-layer features of the decoder between the watermarked image and its distorted images. This alignment encourages the model to extract distortion-invariant representations, thereby improving the reliability of watermark recovery.
The primary contributions of this paper are summarized as follows:

\begin{itemize}
    \item We reveal that the SRD training strategy inherently suffers from overfitting and gradient conflicts. To this end, we reform this training paradigm by proposing a novel meta-learning strategy (Meta-FC).
    \item Meta-FC simulates training on known distortions and testing on ``unknown” distortions within each batch, guiding the model to learn stable and adaptable parameters. Meanwhile, we introduce a feature consistency loss that aligns decoder features between watermarked and distorted images, promoting the learning of distortion-invariant representations.
    \item Meta-FC is a plug-and-play training strategy that can be seamlessly integrated into any existing END-based watermarking model.
    \item Extensive experiments demonstrate that Meta-FC significantly improves the robustness and generalization of different watermarking models under high-intensity, combined, and unknown distortions.
\end{itemize}

%-------------------------------------------------------------------------
\section{Related Work}
\subsection{Deep Learning for Robust Watermarking}
In recent years, with the advancement of deep learning, the DNN-based watermarking frameworks have been proposed \cite{zhu2018hidden, ahmadi2020redmark, tancik2020stegastamp, liu2019novel, jia2021mbrs, ma2022towards, fang2023flow, Wu2023sepmark, zhang2024editguard, liu2025watermarking}. 
HiDDeN \cite{zhu2018hidden} introduced the first end-to-end watermarking model utilizing an END architecture.
ReDMark \cite{ahmadi2020redmark} improved the imperceptibility and robustness of the watermarks by incorporating residual structures within the encoder. StegaStamp \cite{tancik2020stegastamp} achieved robustness against print-and-capture distortions by combining differentiable simulation noise layers and spatial transformation modules, providing a novel approach to handling non-differentiable distortions. TSDL \cite{liu2019novel} employed a two-stage separation architecture to achieve robustness against black-box distortions. MBRS \cite{jia2021mbrs} enhances robustness against JPEG compression by cyclically training on clean images, real JPEG images, and simulated JPEG images across mini-batches. FIN \cite{fang2023flow} proposed a flow-based watermarking framework that uses the invertibility of invertible neural networks (INNs) to enable weight sharing between the encoder and decoder, thereby ensuring tight coupling between the encoder and decoder. Subsequent works \cite{fang2024dero, sun2025end} have further improved model robustness by modifying the internal design of the models, simulating noise layers, and altering the END architecture. These methods adopt the SRD training strategy, in which different distortions are alternated across training batches to achieve robustness against different distortions. However, SRD cannot effectively capture the commonalities between different distortions, making it difficult for the model to learn distortion-invariant representations.

\begin{figure*}[ht]
\centering
\includegraphics[width=0.87\textwidth]{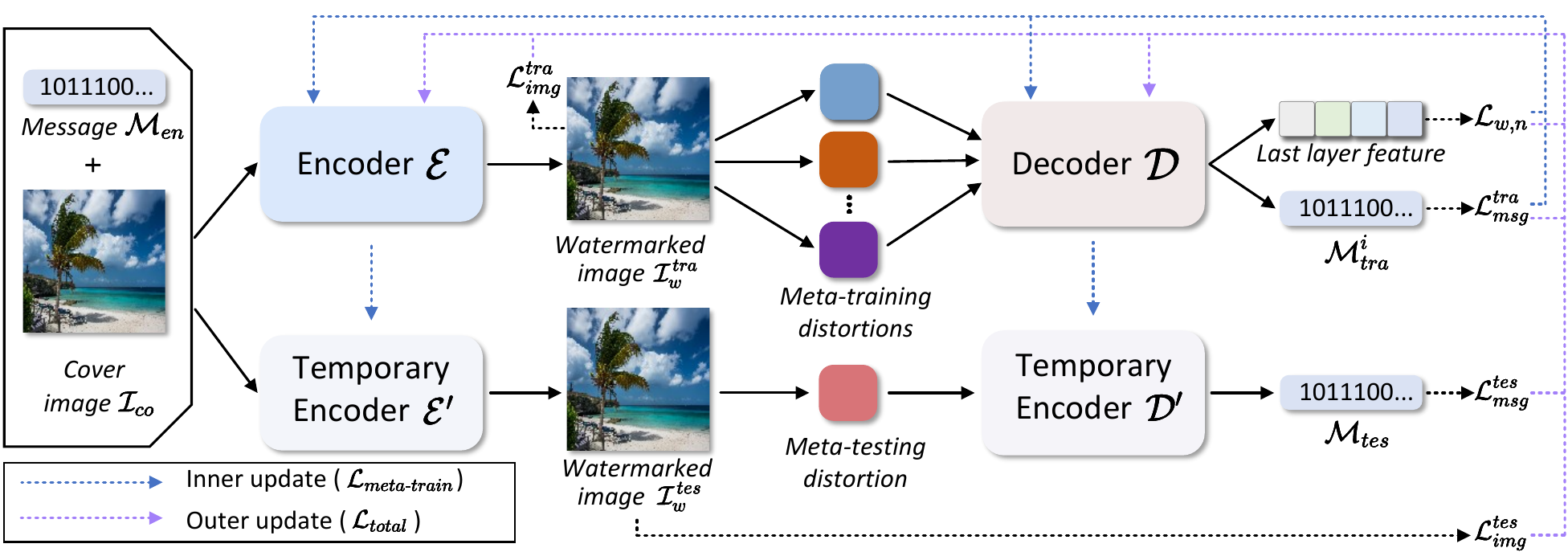}
\caption{The whole training process of our proposed Meta-FC. First, the main encoder and decoder are used to process images under meta-training distortions, yielding the meta-training loss $\mathcal{L}_{meta\text{-}train}$ (composed of $\mathcal{L}_{w,n}$ and $\mathcal{L}^{tra}_{msg}$) and producing temporary encoder and decoder parameters. Next, these temporary parameters are evaluated on the meta-testing distortions to calculate the meta-testing loss $\mathcal{L}_{meta\text{-}test}$ (composed of $\mathcal{L}^{tes}_{msg}$). Subsequently, the image loss $\mathcal{L}_{img}$ (composed of $\mathcal{L}^{tra}_{img}$ and $\mathcal{L}^{tes}_{img}$) is computed based on the watermarked images generated by the main and the temporary encoders. Finally, the main model parameters are updated by minimizing the total loss $\mathcal{L}_{total}$, which consists of $\mathcal{L}_{meta\text{-}train}$, $\mathcal{L}_{meta\text{-}test}$, and $\mathcal{L}_{img}$.}
\label{fig:figure2}
\end{figure*}

\subsection{Meta-learning}
Meta-learning \cite{hsu2018unsupervised, volpi2021continual, zhang2024meta} aims to enhance the model's ability to ``learn how to learn", enabling it to adapt more effectively to new tasks. 
MAML \cite{finn2017model} optimizes the model's initial parameters through two gradient updates, allowing it to quickly fine-tune and adapt to new tasks. 
MLDG \cite{li2018learning} uses meta-learning to simulate domain shifts during training, guiding the model to learn domain-invariant features and significantly enhancing its generalization ability. 
MGAA \cite{yuan2021meta} utilizes meta-learning to bridge the gradient gap between white-box and black-box attacks, improving the transferability of adversarial attacks. 
MLDGG \cite{tian2024mldgg} combines meta-learning with graph neural networks to enhance the model's adaptability in cross-graph domain tasks. 
In fact, these methods essentially leverage meta-learning to mine invariant features across different tasks, improving the model's generalization ability. Inspired by these works, we are naturally motivated to design a meta-learning training strategy for watermarking models that learn invariant features of images under various distortions, thereby improving their generalization ability.

\section{Proposed Meta-FC Method}
\subsection{Key Insight and Intuition}
Existing deep learning-based watermarking methods typically achieve robustness against distortions by applying the SRD training strategy. However, SRD handles each distortion independently, without modeling the underlying commonalities among different distortions, which leads to limited generalization ability.
Inspired by the success of meta-learning in domain generalization, we observe that meta-learning facilitates the extraction of shared representations across multiple domains through iterative meta-train and meta-test. This paradigm can be naturally extended to the modeling of diverse distortions in watermarking.

Building on this insight, we formulate different combinations of distortions as tasks within a meta-learning framework and train the model to continuously adapt and optimize across diverse distortions. This approach gradually discovers a set of robust parameters that are resilient to various perturbations, thereby alleviating the optimization conflicts arising from competing distortion objectives. Furthermore, the meta-testing phase, by simulating ``unknown'' distortions, effectively evaluates the model’s generalization to such novel distortions, thereby improving its robustness against distortions not encountered during training.

\subsection{Meta-FC Pipeline} 
The proposed Meta-FC is a model-agnostic training strategy. As shown in Figure \ref{fig:figure2}, we simulate meta-train with known distortions and meta-test with ``unknown'' distortions to narrow the gap of gradient directions between known and ``unknown'' distortions, improving the generalization of the watermarking model. Specifically, given a noise pool containing $m+1$ distortions, we randomly sample $m$ distortions as meta-training distortions and use the remaining one as the meta-testing distortion in each batch.

\subsubsection{Meta-train.}
The main encoder $\mathcal{E}$ and decoder $\mathcal{D}$ are trained using $m$ randomly selected meta-training distortions, resulting in $m$ decoded messages $\mathcal{M}^{i}_{tra}$ and their corresponding last layer decoder features. The decoding loss $\mathcal{L}_{msg}^{tra}$ is calculated based on $\mathcal{M}^{i}_{tra}$, while the feature consistency loss $\mathcal{L}_{w,n}$ is calculated using the extracted features, which is detailed in Section \textbf{Feature Consistency Loss}. The $\mathcal{L}^{tra}_{msg}$ and $\mathcal{L}_{w,n}$ are aggregated to form the meta-training loss $\mathcal{L}_{meta\text{-}train}$, which is used to update the model parameters and obtain a temporary encoder $\mathcal{E}'$ and decoder $\mathcal{D}'$. This adaptation process across multiple distortions helps mitigate the optimization conflicts introduced by differing distortion objectives. $\mathcal{L}_{meta\text{-}train}$ is defined as follows:

\begin{equation}
\resizebox{0.9\linewidth}{!}{$
\begin{gathered}
\mathcal{L}^{tra}_{msg}
= \sum_{i=1}^{m} \textit{MSE}(\mathcal{M}_{en}, \mathcal{M}^{i}_{tra})
= \sum_{i=1}^{m} \textit{MSE}(\mathcal{M}_{en}, \mathcal{D}(\theta_d, \mathcal{I}^{i}_{tra})), \\[3pt]
\mathcal{L}_{meta\text{-}train}
= \mathcal{L}^{tra}_{msg} + \lambda_f \cdot \mathcal{L}_{w,n},
\end{gathered}
$}
\label{eq1}
\end{equation}
where $\mathcal{M}_{en}$ represents the embedded watermark message and \textit{MSE} represents the mean squared error. $\mathcal{I}^{i}_{tra}$ denotes the watermarked image after the $i$-th distortion in the meta-training distortions, and $\mathcal{D}(\theta_d, \cdot)$ is the decoder with the parameter $\theta_d$. The hyperparameter $\lambda_f$ balances $\mathcal{L}^{tra}_{msg}$ and $\mathcal{L}_{w, n}$, and is set to $0.001$ by default.

\subsubsection{Meta-test.}
After training on multiple meta-training distortions, the sampled meta-testing distortion is used to simulate an ``unknown'' distortion. The $\mathcal{E}'$ and $\mathcal{D}'$ are employed to compute the meta-testing loss $\mathcal{L}_{meta\text{-}test}$, which evaluates the generalization of the model to previously ``unknown'' distortions. $\mathcal{L}_{meta\text{-}test}$ is defined as:

\begin{equation}
\resizebox{0.9\linewidth}{!}{$
\begin{gathered}
\mathcal{L}^{tes}_{msg}
= \textit{MSE}(\mathcal{M}_{en}, \mathcal{M}_{tes})
= \textit{MSE}(\mathcal{M}_{en}, \mathcal{D}'(\theta'_d, \mathcal{I}_{tes})), \\
\mathcal{L}_{meta\text{-}test}
= \mathcal{L}^{tes}_{msg},
\end{gathered}
$}
\label{eq2}
\end{equation}
where $\mathcal{L}^{tes}_{msg}$ is the message loss during the meta-testing phase and $\mathcal{M}_{tes}$ is the message decoded from the distorted image $\mathcal{I}_{tes}$.

Moreover, the watermarked image should maintain the visual quality of $\mathcal{I}_{co}$. To achieve this goal, we also introduce an image loss $\mathcal{L}_{img}$, computed using both main and temporary encoders, as follows:

\begin{equation}
\resizebox{0.9\linewidth}{!}{$
\begin{gathered}
\mathcal{L}^{tra}_{img}
= \textit{MSE}(\mathcal{I}_{co}, \mathcal{I}^{tra}_{w})
= \textit{MSE}(\mathcal{I}_{co}, \mathcal{E}(\theta_e, \mathcal{I}_{co}, \mathcal{M}_{en})), \\
\mathcal{L}^{tes}_{img}
= \textit{MSE}(\mathcal{I}_{co}, \mathcal{I}^{tes}_{w})
= \textit{MSE}(\mathcal{I}_{co}, \mathcal{E}'(\theta'_e, \mathcal{I}_{co}, \mathcal{M}_{en})), \\
\mathcal{L}_{img}
= \mathcal{L}^{tra}_{img} + \mathcal{L}^{tes}_{img},
\end{gathered}
$}
\label{eq3}
\end{equation}
here, $\mathcal{I}^{tra}_{w}$ and $\mathcal{I}^{tes}_{w}$ are the watermarked image generated by the main encoder $\mathcal{E}(\theta_e, \cdot)$ and the temporary encoder $\mathcal{E}^{\prime}(\theta'_e, \cdot)$, respectively.

Ultimately, we aggregated $\mathcal{L}_{meta\text{-}train}$, $\mathcal{L}_{meta\text{-}test}$, and $\mathcal{L}_{img}$ through a weighted combination to update the parameters of the main model, achieving joint optimization of imperceptibility and robustness. The total loss function $\mathcal{L}_{total}$ is defined as:

\begin{equation}
\resizebox{0.9\linewidth}{!}{$
\begin{gathered}
\mathcal{L}_{total}
= \lambda_1 \cdot \frac{\mathcal{L}_{meta\text{-}train} + \mathcal{L}_{meta\text{-}test}}{m+1}
+ \lambda_2 \cdot \frac{\mathcal{L}_{img}}{2},
\end{gathered}
$}
\label{eq4}
\end{equation}
where $\lambda_1$ and $\lambda_2$ are hyperparameters that balance the robustness and imperceptibility.

Throughout the training process, the weighting factors $\lambda_1$ and $\lambda_2$ are dynamically adjusted, similar to \cite{xu2025invismark}. In the early training stages, the model focuses on learning robust watermark decoding. Therefore, $\lambda_1$ is initialized to $5$, and $\lambda_2$ is initialized to a relatively low value of $1$. This design helps stabilize optimization since the embedded watermark signal is inherently weaker than the image content. As training progresses and the model achieves stable watermark recovery under various distortions, $\lambda_1$ is gradually decreased to its predefined minimum value of $1$, while $\lambda_2$ is progressively increased to its maximum value of $15$ to further improve the visual quality of the watermarked images.

\begin{algorithm}
\caption{\textit{Meta-FC}}
\label{alg: Meta-FC Algorithm}
\begin{algorithmic}[1]
\REQUIRE cover images \(\mathcal{I}_{co}\), watermark message \(\mathcal{M}_{en}\), watermark encoder \(\mathcal{E}\), watermark decoder \(\mathcal{D}\), noise pool $\mathcal{N}$ including \(m+1\) distortions.
\ENSURE best model parameters.

\STATE Initialization;
\FOR{$k \in [0, maxiter]$}
    \STATE Randomly sample $m$ distortions from the noise pool;\STATE \textit{// Meta-Train}
    \STATE $\mathcal{I}^{tra}_{w} \leftarrow \mathcal{E}(\mathcal{I}_{co}, \mathcal{M}_{en})$;
    
    \FOR{$i \in [1, m]$}
        \STATE $\mathcal{I}^{i}_{tra} \leftarrow \mathcal{N}(\mathcal{I}_{w}^{tra})$;
        \STATE $\mathcal{M}^{i}_{tra} \leftarrow \mathcal{D}(\mathcal{I}^{i}_{tra})$;
    \ENDFOR
    \STATE Compute $\mathcal{L}_{w, n}$ with Eq.(\ref{eq5}$\&$\ref{eq6});
    \STATE Compute $\mathcal{L}_{meta\text{-}train}$ and $\mathcal{L}^{tra}_{msg}$ with Eq.(\ref{eq1});
    \STATE Inner update and get temporary model \(\mathcal{E}'\) and \(\mathcal{D}'\);
    \STATE Sample the remaining distortion from the noise pool;\STATE \textit{// Meta-Test}
    \STATE $\mathcal{I}^{tes}_{w} \leftarrow \mathcal{E}'(\mathcal{I}_{co}, \mathcal{M}_{en})$;
    \STATE $\mathcal{I}_{tes} \leftarrow \mathcal{N}(\mathcal{I}^{tes}_{w})$;
    \STATE $\mathcal{M}_{tes} \leftarrow \mathcal{D}'(\mathcal{I}_{tes})$;
    \STATE Compute $\mathcal{L}_{meta\text{-}test}$ with Eq.(\ref{eq2});
    \STATE \textit{// Outer update}
    \STATE Compute $\mathcal{L}_{img}$ and $\mathcal{L}_{total}$ with Eq.(\ref{eq3}$\&$\ref{eq4});
    \STATE Update \(\theta_e\) and \(\theta_d\);
\ENDFOR
\STATE \textbf{Return:} best model parameters.
\end{algorithmic}
\end{algorithm}

\begin{figure*}[!t]
\centering
\includegraphics[width=0.96\textwidth]{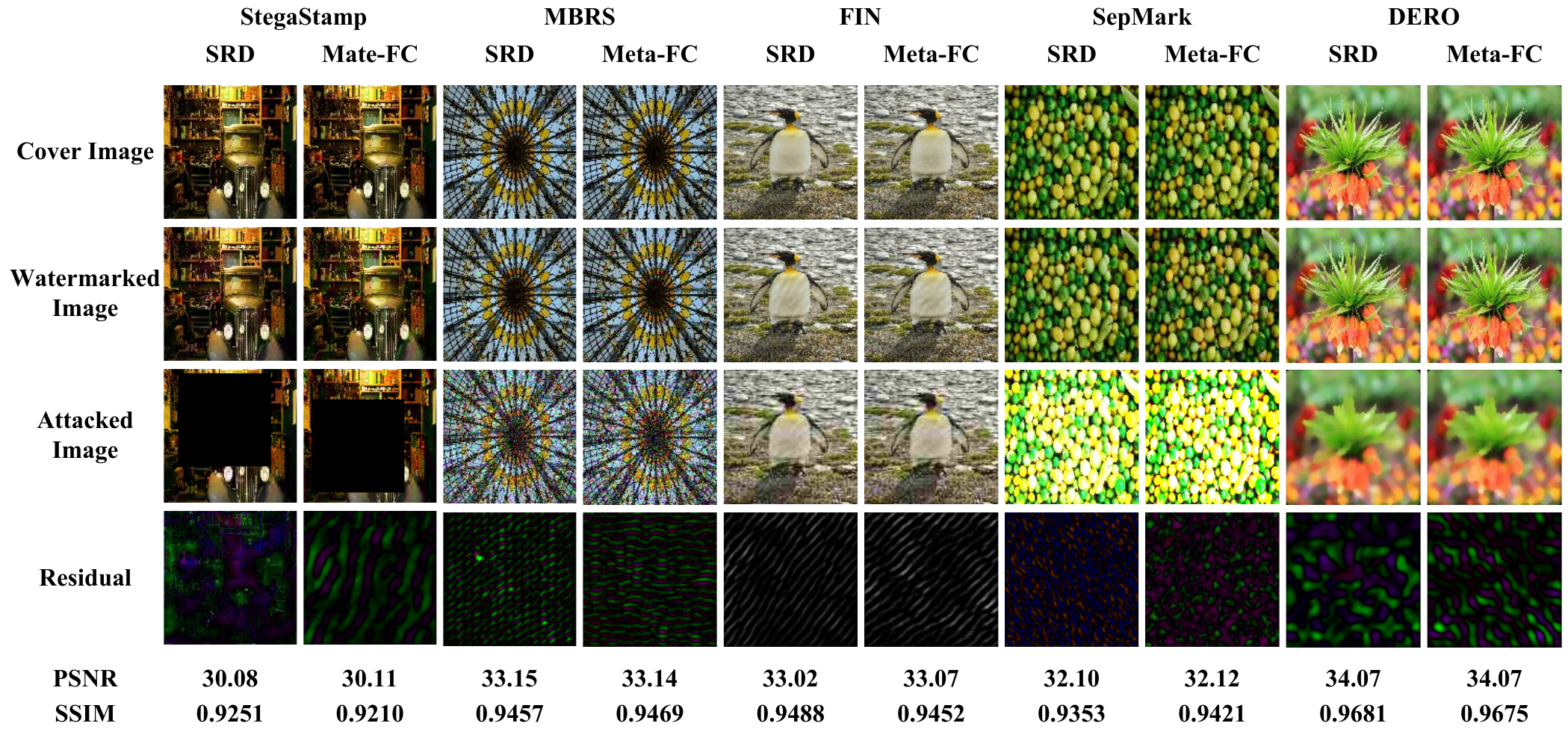}
\caption{Result of the visual quality of SRD and our method under different models. The first row presents the cover image, followed by the watermarked images in the second row. The third row shows the watermarked images subjected to various distortions. The fourth row illustrates the residuals, which represent the difference between the watermarked and cover images and are magnified by a factor of 5 to enhance visibility. The final two rows report the PSNR(dB) and SSIM values of each model, respectively, where consistent visual quality is maintained across training methods for the same model.}
\label{fig:figure3}
\end{figure*}

\subsection{Feature Consistency Loss}
Although meta-learning helps select parameters that are robust across diverse distortions, it remains limited in improving the representational capacity of the model itself. 
To enhance distortion-invariant representations, we introduce a feature consistency loss that aligns the decoder features between $\mathcal{I}^{tra}_{w}$ and $\mathcal{I}^{i}_{tra}$. The core intuition is that the decoder features extracted from $\mathcal{I}^{tra}_{w}$ typically preserve more complete and reliable watermark information. By encouraging the features from $\mathcal{I}^{i}_{tra}$ to match those of $\mathcal{I}^{tra}_{w}$, the model learns to extract watermark-relevant representations that are more resilient to distortions.

Let $f_{w}$ denote the last layer decoder feature vector extracted from $\mathcal{I}^{tra}_{w}$, and let $f_{no}^{i}$ denote the last layer decoder feature extracted from $\mathcal{I}^{i}_{tra}$. These feature vectors are normalized as follows:

\begin{equation}
\begin{gathered}
\bar{f}_{w} = \frac{f_{w}}{\|f_{w}\|_2}, \qquad
\bar{f}_{no}^{i} = \frac{f_{no}^{i}}{\|f_{no}^{i}\|_2}.
\end{gathered}
\label{eq5}
\end{equation}
Using $\bar{f}_{w}$ as anchor, we quantify the distance between $\bar{f}_{w}$ and $\bar{f}_{no}^{i}$ through cosine similarity:

\begin{equation}
\resizebox{0.9\linewidth}{!}{$
\begin{gathered}
\mathcal{L}_{w,n}
= \sum_{i=1}^{m} \left( 1 - \cos(\bar{f}_{w}, \bar{f}_{no}^{i}) \right) 
= \sum_{i=1}^{m} \left( 1 - \langle \bar{f}_{w}, \bar{f}_{no}^{i} \rangle \right).
\end{gathered}
$}
\label{eq6}
\end{equation}
By minimizing this loss, the decoded features from all distortions are encouraged to converge toward a consistent watermark representation, thereby enhancing the model's robustness to both high-intensity and combined distortions.

\section{Experiments}
\subsection{Experiments Settings}
\subsubsection{Setting Details.} In this paper, all experiments are implemented using PyTorch \cite{Collobert2011torch7} and executed on an NVIDIA GeForce RTX 3090 GPU. The models are trained in DIV2K \cite{agustsson2017ntire} and use the DIV2K test set together with $2,000$ additional images randomly sampled from COCO \cite{lin2014microsoft} and ImageNet \cite{deng2009imagenet} to evaluate the generalization performance. All images are resized to $128\times128$, and the watermark message length is fixed to $64$ bits. The learning rate and batch size are set to $0.001$ and $16$. 

\subsubsection{Baselines.} To evaluate the robustness and generalization of Meta-FC, we employ five representative watermarking models with diverse architectures, including StegaStamp \cite{tancik2020stegastamp}, MBRS \cite{jia2021mbrs}, FIN \cite{fang2023flow}, SepMark \cite{Wu2023sepmark}, and DERO \cite{fang2024dero}. For each model, we retain the original encoder and decoder architectures and train them using both our proposed Meta-FC and the baseline SRD approach to ensure a fair comparison.

\subsubsection{Metrics.} We evaluate watermark robustness using average bit accuracy (ACC), which measures the proportion of correctly recovered bits from the extracted watermarks. To assess visual quality, we report peak signal-to-noise ratio (PSNR) and structural similarity (SSIM) as metrics. To eliminate statistical bias, we repeated the experiments with different random seeds and reported the mean result.

\begin{table*}[htbp]
  \centering
  \setlength{\tabcolsep}{9pt}
  \caption{Comparison of ACC (\%) under high-intensity distortions. To ensure a fair comparison, the embedding strength of each baseline model is adjusted for both SRD and Meta-FC, such that the resulting PSNR values remain consistent across methods.}
  \label{tab:table1}
  \renewcommand{\arraystretch}{0.9}
  \resizebox{0.9\linewidth}{!}{
  \begin{tabular}{ccc|cccccccccc}
    \toprule
    \textbf{Datasets} & \textbf{Models} & \textbf{Strategy} & \textbf{Identity} & 
    \begin{tabular}[c]{@{}c@{}}\textbf{GB}\\ ($\sigma=6$)\end{tabular} & 
    \begin{tabular}[c]{@{}c@{}}\textbf{SPN}\\ ($\sigma=0.15$)\end{tabular} & 
    \begin{tabular}[c]{@{}c@{}}\textbf{Crop}\\ ($p=0.7$)\end{tabular} & 
    \begin{tabular}[c]{@{}c@{}}\textbf{JPEG}\\ ($\mathcal{Q}=30$)\end{tabular} & 
    \begin{tabular}[c]{@{}c@{}}\textbf{MB}\\ ($w=7$)\end{tabular} & 
    \begin{tabular}[c]{@{}c@{}}\textbf{GN}\\ ($\sigma=0.08$)\end{tabular} & 
    \begin{tabular}[c]{@{}c@{}}\textbf{Bright}\\ ($f=4$)\end{tabular} & 
    \begin{tabular}[c]{@{}c@{}}\textbf{Dropout}\\ ($p=0.7$)\end{tabular} &
    \textbf{Avg.} \\
    \midrule

    % ==================== DIV2K ====================
    \multirow{10}{*}{\rotatebox{90}{DIV2K}} 
      & \multirow{2}{*}{StegaStamp} 
        & SRD & \textbf{100} & \textbf{100} & 98.07 & 86.56 & 95.84 & 99.97 & 97.89 & 77.70 & 94.33 & 94.48 \\
      & & \cellcolor{table_col}Meta-FC 
        & \cellcolor{table_col}\textbf{100} 
        & \cellcolor{table_col}\textbf{100} 
        & \cellcolor{table_col}\textbf{99.07} 
        & \cellcolor{table_col}\textbf{90.00} 
        & \cellcolor{table_col}\textbf{97.42} 
        & \cellcolor{table_col}\textbf{99.99} 
        & \cellcolor{table_col}\textbf{98.25} 
        & \cellcolor{table_col}\textbf{80.84} 
        & \cellcolor{table_col}\textbf{94.56} & \cellcolor{table_col}\textbf{95.57} \\
      \cmidrule(l){2-13}

      & \multirow{2}{*}{MBRS} 
        & SRD & \textbf{100} & 99.02 & 97.46 & 91.94 & 83.95 & 90.82 & 93.98 & \textbf{94.25} & \textbf{99.76} & 94.58 \\
      & & \cellcolor{table_col}Meta-FC 
        & \cellcolor{table_col}\textbf{100} 
        & \cellcolor{table_col}\textbf{99.33} 
        & \cellcolor{table_col}\textbf{98.44} 
        & \cellcolor{table_col}\textbf{95.78} 
        & \cellcolor{table_col}\textbf{86.79} 
        & \cellcolor{table_col}\textbf{95.25} 
        & \cellcolor{table_col}\textbf{96.01} 
        & \cellcolor{table_col}93.93 
        & \cellcolor{table_col}99.75 & \cellcolor{table_col}\textbf{96.14} \\
      \cmidrule(l){2-13}

      & \multirow{2}{*}{FIN} 
        & SRD & 99.97 & 96.25 & 97.27 & 81.16 & 99.04 & 90.15 & 93.89 & 88.73 & 83.29 & 92.19 \\
      & & \cellcolor{table_col}Meta-FC 
        & \cellcolor{table_col}\textbf{100} 
        & \cellcolor{table_col}\textbf{97.16} 
        & \cellcolor{table_col}\textbf{98.27} 
        & \cellcolor{table_col}\textbf{83.15} 
        & \cellcolor{table_col}\textbf{99.89} 
        & \cellcolor{table_col}\textbf{90.18} 
        & \cellcolor{table_col}\textbf{95.70} 
        & \cellcolor{table_col}\textbf{89.32} 
        & \cellcolor{table_col}\textbf{85.58} & \cellcolor{table_col}\textbf{93.25} \\
      \cmidrule(l){2-13}

      & \multirow{2}{*}{SepMark} 
        & SRD & 99.95 & 98.54 & 94.60 & 99.27 & 79.05 & 95.59 & 91.62 & 96.42 & 97.75 & 94.75 \\
      & & \cellcolor{table_col}Meta-FC 
        & \cellcolor{table_col}\textbf{100} 
        & \cellcolor{table_col}\textbf{99.85} 
        & \cellcolor{table_col}\textbf{98.26} 
        & \cellcolor{table_col}\textbf{99.32} 
        & \cellcolor{table_col}\textbf{86.57} 
        & \cellcolor{table_col}\textbf{99.72} 
        & \cellcolor{table_col}\textbf{96.70} 
        & \cellcolor{table_col}\textbf{97.20} 
        & \cellcolor{table_col}\textbf{97.90} & \cellcolor{table_col}\textbf{97.28} \\
      \cmidrule(l){2-13}

      & \multirow{2}{*}{DERO} 
        & SRD & 99.96 & 99.97 & 97.79 & 95.72 & 89.26 & 99.89 & 92.16 & 92.29 & 96.24 & 95.92 \\
      & & \cellcolor{table_col}Meta-FC 
        & \cellcolor{table_col}\textbf{100} 
        & \cellcolor{table_col}\textbf{100} 
        & \cellcolor{table_col}\textbf{99.61} 
        & \cellcolor{table_col}\textbf{97.44} 
        & \cellcolor{table_col}\textbf{91.78} 
        & \cellcolor{table_col}\textbf{100} 
        & \cellcolor{table_col}\textbf{95.04} 
        & \cellcolor{table_col}\textbf{92.86} 
        & \cellcolor{table_col}\textbf{98.10} & \cellcolor{table_col}\textbf{97.20} \\
    \midrule

    % ==================== COCO ====================
    \multirow{10}{*}{\rotatebox{90}{COCO}} 
      & \multirow{2}{*}{StegaStamp} 
        & SRD & \textbf{100} & \textbf{100} & 98.47 & 86.88 & 96.66 & 99.99 & 98.44 & 78.45 & 94.48 & 94.82 \\
      & & \cellcolor{table_col}Meta-FC 
        & \cellcolor{table_col}\textbf{100} 
        & \cellcolor{table_col}\textbf{100} 
        & \cellcolor{table_col}\textbf{99.20} 
        & \cellcolor{table_col}\textbf{90.09} 
        & \cellcolor{table_col}\textbf{97.93} 
        & \cellcolor{table_col}\textbf{100} 
        & \cellcolor{table_col}\textbf{98.89} 
        & \cellcolor{table_col}\textbf{81.04} 
        & \cellcolor{table_col}\textbf{94.80} & \cellcolor{table_col}\textbf{95.77} \\
      \cmidrule(l){2-13}

      & \multirow{2}{*}{MBRS} 
        & SRD & \textbf{100} & 99.30 & 97.89 & 92.26 & 87.73 & 93.58 & 94.32 & 91.22 & \textbf{99.89} & 95.13 \\
      & & \cellcolor{table_col}Meta-FC 
        & \cellcolor{table_col}\textbf{100} 
        & \cellcolor{table_col}\textbf{99.77} 
        & \cellcolor{table_col}\textbf{98.51} 
        & \cellcolor{table_col}\textbf{96.42} 
        & \cellcolor{table_col}\textbf{90.16} 
        & \cellcolor{table_col}\textbf{97.71} 
        & \cellcolor{table_col}\textbf{97.02} 
        & \cellcolor{table_col}\textbf{91.60} 
        & \cellcolor{table_col}99.87 & \cellcolor{table_col}\textbf{96.78} \\
      \cmidrule(l){2-13}

      & \multirow{2}{*}{FIN} 
        & SRD & \textbf{100} & 96.35 & 95.98 & 82.13 & 99.16 & \textbf{91.59} & 91.95 & 85.92 & 82.14 & 91.69 \\
      & & \cellcolor{table_col}Meta-FC 
        & \cellcolor{table_col}\textbf{100} 
        & \cellcolor{table_col}\textbf{97.43} 
        & \cellcolor{table_col}\textbf{98.57} 
        & \cellcolor{table_col}\textbf{83.82} 
        & \cellcolor{table_col}\textbf{99.91} 
        & \cellcolor{table_col}91.21 
        & \cellcolor{table_col}\textbf{94.55} 
        & \cellcolor{table_col}\textbf{86.29} 
        & \cellcolor{table_col}\textbf{88.96} & \cellcolor{table_col}\textbf{93.42} \\
      \cmidrule(l){2-13}

      & \multirow{2}{*}{SepMark} 
        & SRD & \textbf{100} & 98.87 & 94.90 & 99.56 & 81.36 & 97.50 & 92.32 & 96.65 & 98.91 & 95.56 \\
      & & \cellcolor{table_col}Meta-FC 
        & \cellcolor{table_col}\textbf{100} 
        & \cellcolor{table_col}\textbf{99.98} 
        & \cellcolor{table_col}\textbf{98.45} 
        & \cellcolor{table_col}\textbf{99.63} 
        & \cellcolor{table_col}\textbf{86.96} 
        & \cellcolor{table_col}\textbf{99.94} 
        & \cellcolor{table_col}\textbf{96.86} 
        & \cellcolor{table_col}\textbf{97.42} 
        & \cellcolor{table_col}\textbf{99.26} & \cellcolor{table_col}\textbf{97.61} \\
      \cmidrule(l){2-13}

      & \multirow{2}{*}{DERO} 
        & SRD & 99.99 & 99.99 & 98.08 & 96.40 & 88.66 & 99.95 & 92.98 & 88.94 & 96.84 & 95.76 \\
      & & \cellcolor{table_col}Meta-FC 
        & \cellcolor{table_col}\textbf{100} 
        & \cellcolor{table_col}\textbf{100} 
        & \cellcolor{table_col}\textbf{99.70} 
        & \cellcolor{table_col}\textbf{96.98} 
        & \cellcolor{table_col}\textbf{89.93} 
        & \cellcolor{table_col}\textbf{100} 
        & \cellcolor{table_col}\textbf{95.13} 
        & \cellcolor{table_col}\textbf{89.25} 
        & \cellcolor{table_col}\textbf{98.59} & \cellcolor{table_col}\textbf{96.62} \\
    \midrule

    % ==================== ImageNet ====================
    \multirow{10}{*}{\rotatebox{90}{ImageNet}} 
      & \multirow{2}{*}{StegaStamp} 
        & SRD & \textbf{100} & \textbf{100} & 98.66 & 86.87 & 96.56 & \textbf{100} & 98.05 & 78.52 & 94.47 & 94.79 \\
      & & \cellcolor{table_col}Meta-FC 
        & \cellcolor{table_col}\textbf{100} 
        & \cellcolor{table_col}\textbf{100} 
        & \cellcolor{table_col}\textbf{99.38} 
        & \cellcolor{table_col}\textbf{90.12} 
        & \cellcolor{table_col}\textbf{97.67} 
        & \cellcolor{table_col}\textbf{100} 
        & \cellcolor{table_col}\textbf{98.76} 
        & \cellcolor{table_col}\textbf{81.62} 
        & \cellcolor{table_col}\textbf{94.75} & \cellcolor{table_col}\textbf{95.81} \\
      \cmidrule(l){2-13}

      & \multirow{2}{*}{MBRS} 
        & SRD & \textbf{100} & 97.99 & 97.85 & 92.37 & 86.27 & 92.14 & 94.14 & \textbf{95.58} & 99.84 & 95.13 \\
      & & \cellcolor{table_col}Meta-FC 
        & \cellcolor{table_col}\textbf{100} 
        & \cellcolor{table_col}\textbf{98.71} 
        & \cellcolor{table_col}\textbf{98.59} 
        & \cellcolor{table_col}\textbf{96.64} 
        & \cellcolor{table_col}\textbf{89.51} 
        & \cellcolor{table_col}\textbf{96.99} 
        & \cellcolor{table_col}\textbf{97.02} 
        & \cellcolor{table_col}95.48 
        & \cellcolor{table_col}\textbf{99.97} & \cellcolor{table_col}\textbf{96.99} \\
      \cmidrule(l){2-13}

      & \multirow{2}{*}{FIN} 
        & SRD & \textbf{100} & 96.29 & 97.27 & 81.58 & 99.39 & \textbf{89.62} & 91.52 & 88.88 & 85.95 & 92.28 \\
      & & \cellcolor{table_col}Meta-FC 
        & \cellcolor{table_col}\textbf{100} 
        & \cellcolor{table_col}\textbf{97.48} 
        & \cellcolor{table_col}\textbf{98.27} 
        & \cellcolor{table_col}\textbf{83.54} 
        & \cellcolor{table_col}\textbf{99.86} 
        & \cellcolor{table_col}89.49 
        & \cellcolor{table_col}\textbf{94.76} 
        & \cellcolor{table_col}\textbf{89.08} 
        & \cellcolor{table_col}\textbf{96.65} & \cellcolor{table_col}\textbf{94.35} \\
      \cmidrule(l){2-13}

      & \multirow{2}{*}{SepMark} 
        & SRD & 99.96 & 98.77 & 94.90 & 99.59 & 80.85 & 96.72 & 91.80 & 94.33 & 98.78 & 95.08 \\
      & & \cellcolor{table_col}Meta-FC 
        & \cellcolor{table_col}\textbf{100} 
        & \cellcolor{table_col}\textbf{100} 
        & \cellcolor{table_col}\textbf{98.45} 
        & \cellcolor{table_col}\textbf{99.64} 
        & \cellcolor{table_col}\textbf{87.92} 
        & \cellcolor{table_col}\textbf{99.96} 
        & \cellcolor{table_col}\textbf{96.99} 
        & \cellcolor{table_col}\textbf{98.03} 
        & \cellcolor{table_col}\textbf{99.36} & \cellcolor{table_col}\textbf{97.82} \\
      \cmidrule(l){2-13}

      & \multirow{2}{*}{DERO} 
        & SRD & \textbf{100} & 99.97 & 97.79 & 96.23 & 90.38 & 99.99 & 92.91 & 91.81 & 97.09 & 96.24 \\
      & & \cellcolor{table_col}Meta-FC 
        & \cellcolor{table_col}\textbf{100} 
        & \cellcolor{table_col}\textbf{100} 
        & \cellcolor{table_col}\textbf{99.61} 
        & \cellcolor{table_col}\textbf{96.91} 
        & \cellcolor{table_col}\textbf{92.03} 
        & \cellcolor{table_col}\textbf{100} 
        & \cellcolor{table_col}\textbf{95.34} 
        & \cellcolor{table_col}\textbf{92.56} 
        & \cellcolor{table_col}\textbf{99.04} & \cellcolor{table_col}\textbf{97.28} \\
    \bottomrule
  \end{tabular}
  }
  \vspace{-6pt}
\end{table*}

\begin{table*}[htbp]
  \centering
  \setlength{\tabcolsep}{7pt}
  \caption{Comparison of ACC (\%) under combined distortions. PSNR is kept consistent across training methods for each baseline, following the same setup as in the high-intensity distortion evaluation.}
  \label{tab:table2}
  \renewcommand{\arraystretch}{0.9}
  % \scalebox{0.65}{
  \resizebox{0.9\linewidth}{!}{
  \begin{tabular}{ccc|cccccccc}
    \toprule
    \textbf{Datasets} & \textbf{Models} & \textbf{Strategy} &
    \textbf{MB\&JPEG} & \textbf{MB\&SPN} & \textbf{MB\&GN} &
    \textbf{GN\&SPN} & \textbf{GN\&JPEG} &
    \begin{tabular}[c]{@{}c@{}}\textbf{MB\&SPN}\\ \textbf{\&GN}\end{tabular} &
    \begin{tabular}[c]{@{}c@{}}\textbf{MB\&JPEG}\\ \textbf{\&Crop}\end{tabular} &
    \textbf{Avg.} \\
    \midrule

    % ==================== DIV2K ====================
    \multirow{10}{*}{\rotatebox{90}{DIV2K}}
      & \multirow{2}{*}{StegaStamp}
        & SRD & 99.13 & \textbf{99.81} & 98.54 & 97.49 & 97.04 & 94.78 & 97.21 & 97.71 \\
      & & \cellcolor{table_col}Meta-FC
        & \cellcolor{table_col}\textbf{99.78}
        & \cellcolor{table_col}\textbf{99.81}
        & \cellcolor{table_col}\textbf{99.01}
        & \cellcolor{table_col}\textbf{98.50}
        & \cellcolor{table_col}\textbf{98.26}
        & \cellcolor{table_col}\textbf{95.52}
        & \cellcolor{table_col}\textbf{97.89}
        & \cellcolor{table_col}\textbf{98.40} \\
      \cmidrule(l){2-11}

      & \multirow{2}{*}{MBRS}
        & SRD & 71.47 & 90.24 & 72.09 & 94.11 & 75.98 & 65.98 & 69.81 & 77.10 \\
      & & \cellcolor{table_col}Meta-FC
        & \cellcolor{table_col}\textbf{81.53}
        & \cellcolor{table_col}\textbf{98.88}
        & \cellcolor{table_col}\textbf{82.70}
        & \cellcolor{table_col}\textbf{96.31}
        & \cellcolor{table_col}\textbf{80.21}
        & \cellcolor{table_col}\textbf{74.92}
        & \cellcolor{table_col}\textbf{79.31}
        & \cellcolor{table_col}\textbf{84.84} \\
      \cmidrule(l){2-11}

      & \multirow{2}{*}{FIN}
        & SRD & 95.98 & 91.23 & \textbf{80.13} & 93.38 & 93.25 & 75.59 & 92.56 & 88.87 \\
      & & \cellcolor{table_col}Meta-FC
        & \cellcolor{table_col}\textbf{97.01}
        & \cellcolor{table_col}\textbf{92.32}
        & \cellcolor{table_col}80.06
        & \cellcolor{table_col}\textbf{95.70}
        & \cellcolor{table_col}\textbf{95.54}
        & \cellcolor{table_col}\textbf{76.69}
        & \cellcolor{table_col}\textbf{95.38}
        & \cellcolor{table_col}\textbf{90.38} \\
      \cmidrule(l){2-11}

      & \multirow{2}{*}{SepMark}
        & SRD & 78.29 & 87.94 & 80.32 & 92.12 & 79.23 & 73.70 & 76.40 & 81.14 \\
      & & \cellcolor{table_col}Meta-FC
        & \cellcolor{table_col}\textbf{94.35}
        & \cellcolor{table_col}\textbf{98.52}
        & \cellcolor{table_col}\textbf{94.29}
        & \cellcolor{table_col}\textbf{96.58}
        & \cellcolor{table_col}\textbf{91.00}
        & \cellcolor{table_col}\textbf{89.39}
        & \cellcolor{table_col}\textbf{92.85}
        & \cellcolor{table_col}\textbf{93.85} \\
      \cmidrule(l){2-11}

      & \multirow{2}{*}{DERO}
        & SRD & 97.49 & 99.38 & 94.48 & 93.31 & 92.14 & 90.64 & 96.52 & 94.85 \\
      & & \cellcolor{table_col}Meta-FC
        & \cellcolor{table_col}\textbf{98.23}
        & \cellcolor{table_col}\textbf{99.97}
        & \cellcolor{table_col}\textbf{95.48}
        & \cellcolor{table_col}\textbf{96.08}
        & \cellcolor{table_col}\textbf{93.75}
        & \cellcolor{table_col}\textbf{92.38}
        & \cellcolor{table_col}\textbf{97.53}
        & \cellcolor{table_col}\textbf{96.20} \\
    \midrule

    % ==================== COCO ====================
    \multirow{10}{*}{\rotatebox{90}{COCO}}
      & \multirow{2}{*}{StegaStamp}
        & SRD & 99.74 & 99.90 & 98.76 & 97.55 & 97.42 & 95.99 & 98.33 & 98.24 \\
      & & \cellcolor{table_col}Meta-FC
        & \cellcolor{table_col}\textbf{99.90}
        & \cellcolor{table_col}\textbf{99.94}
        & \cellcolor{table_col}\textbf{99.24}
        & \cellcolor{table_col}\textbf{98.43}
        & \cellcolor{table_col}\textbf{98.43}
        & \cellcolor{table_col}\textbf{96.98}
        & \cellcolor{table_col}\textbf{98.74}
        & \cellcolor{table_col}\textbf{98.81} \\
      \cmidrule(l){2-11}

      & \multirow{2}{*}{MBRS}
        & SRD & 73.18 & 90.18 & 72.93 & 93.86 & 76.18 & 67.49 & 70.85 & 77.81 \\
      & & \cellcolor{table_col}Meta-FC
        & \cellcolor{table_col}\textbf{84.18}
        & \cellcolor{table_col}\textbf{99.23}
        & \cellcolor{table_col}\textbf{82.57}
        & \cellcolor{table_col}\textbf{96.51}
        & \cellcolor{table_col}\textbf{79.71}
        & \cellcolor{table_col}\textbf{75.28}
        & \cellcolor{table_col}\textbf{82.44}
        & \cellcolor{table_col}\textbf{85.70} \\
      \cmidrule(l){2-11}

      & \multirow{2}{*}{FIN}
        & SRD & 96.18 & 90.86 & 78.28 & 92.62 & 93.83 & 73.65 & 92.77 & 88.31 \\
      & & \cellcolor{table_col}Meta-FC
        & \cellcolor{table_col}\textbf{97.78}
        & \cellcolor{table_col}\textbf{92.20}
        & \cellcolor{table_col}\textbf{78.35}
        & \cellcolor{table_col}\textbf{94.88}
        & \cellcolor{table_col}\textbf{95.14}
        & \cellcolor{table_col}\textbf{75.39}
        & \cellcolor{table_col}\textbf{95.89}
        & \cellcolor{table_col}\textbf{89.95} \\
      \cmidrule(l){2-11}

      & \multirow{2}{*}{SepMark}
        & SRD & 79.47 & 87.79 & 80.00 & 91.75 & 79.28 & 73.39 & 77.91 & 81.37 \\
      & & \cellcolor{table_col}Meta-FC
        & \cellcolor{table_col}\textbf{94.46}
        & \cellcolor{table_col}\textbf{98.64}
        & \cellcolor{table_col}\textbf{95.20}
        & \cellcolor{table_col}\textbf{96.70}
        & \cellcolor{table_col}\textbf{90.81}
        & \cellcolor{table_col}\textbf{90.10}
        & \cellcolor{table_col}\textbf{92.84}
        & \cellcolor{table_col}\textbf{94.11} \\
      \cmidrule(l){2-11}

      & \multirow{2}{*}{DERO}
        & SRD & 97.69 & 99.65 & 94.60 & 93.46 & 91.98 & 90.65 & 96.48 & 94.93 \\
      & & \cellcolor{table_col}Meta-FC
        & \cellcolor{table_col}\textbf{98.05}
        & \cellcolor{table_col}\textbf{99.99}
        & \cellcolor{table_col}\textbf{95.99}
        & \cellcolor{table_col}\textbf{95.65}
        & \cellcolor{table_col}\textbf{93.39}
        & \cellcolor{table_col}\textbf{92.51}
        & \cellcolor{table_col}\textbf{97.03}
        & \cellcolor{table_col}\textbf{96.09} \\
    \midrule

    % ==================== ImageNet ====================
    \multirow{10}{*}{\rotatebox{90}{ImageNet}}
      & \multirow{2}{*}{StegaStamp}
        & SRD & 99.80 & \textbf{99.87} & 98.97 & 97.58 & 97.38 & 95.89 & 98.29 & 98.25 \\
      & & \cellcolor{table_col}Meta-FC
        & \cellcolor{table_col}\textbf{99.86}
        & \cellcolor{table_col}\textbf{99.87}
        & \cellcolor{table_col}\textbf{99.32}
        & \cellcolor{table_col}\textbf{98.60}
        & \cellcolor{table_col}\textbf{98.45}
        & \cellcolor{table_col}\textbf{97.03}
        & \cellcolor{table_col}\textbf{99.05}
        & \cellcolor{table_col}\textbf{98.88} \\
      \cmidrule(l){2-11}

      & \multirow{2}{*}{MBRS}
        & SRD & 72.64 & 90.17 & 72.66 & 94.10 & 75.82 & 67.80 & 70.58 & 77.68 \\
      & & \cellcolor{table_col}Meta-FC
        & \cellcolor{table_col}\textbf{85.41}
        & \cellcolor{table_col}\textbf{99.25}
        & \cellcolor{table_col}\textbf{82.40}
        & \cellcolor{table_col}\textbf{96.57}
        & \cellcolor{table_col}\textbf{80.52}
        & \cellcolor{table_col}\textbf{75.07}
        & \cellcolor{table_col}\textbf{81.44}
        & \cellcolor{table_col}\textbf{85.81} \\
      \cmidrule(l){2-11}

      & \multirow{2}{*}{FIN}
        & SRD & 96.47 & 90.18 & \textbf{78.52} & 92.17 & 93.53 & 73.73 & 92.19 & 88.11 \\
      & & \cellcolor{table_col}Meta-FC
        & \cellcolor{table_col}\textbf{97.26}
        & \cellcolor{table_col}\textbf{91.48}
        & \cellcolor{table_col}78.48
        & \cellcolor{table_col}\textbf{94.79}
        & \cellcolor{table_col}\textbf{95.40}
        & \cellcolor{table_col}\textbf{75.92}
        & \cellcolor{table_col}\textbf{95.34}
        & \cellcolor{table_col}\textbf{89.81} \\
      \cmidrule(l){2-11}

      & \multirow{2}{*}{SepMark}
        & SRD & 78.96 & 87.79 & 79.66 & 91.67 & 79.27 & 73.41 & 77.73 & 81.21 \\
      & & \cellcolor{table_col}Meta-FC
        & \cellcolor{table_col}\textbf{94.83}
        & \cellcolor{table_col}\textbf{98.66}
        & \cellcolor{table_col}\textbf{94.79}
        & \cellcolor{table_col}\textbf{96.61}
        & \cellcolor{table_col}\textbf{90.77}
        & \cellcolor{table_col}\textbf{89.70}
        & \cellcolor{table_col}\textbf{93.58}
        & \cellcolor{table_col}\textbf{94.13} \\
      \cmidrule(l){2-11}

      & \multirow{2}{*}{DERO}
        & SRD & 98.02 & 99.68 & 94.62 & 93.35 & 92.17 & 90.66 & 96.94 & 95.06 \\
      & & \cellcolor{table_col}Meta-FC
        & \cellcolor{table_col}\textbf{98.38}
        & \cellcolor{table_col}\textbf{99.99}
        & \cellcolor{table_col}\textbf{96.06}
        & \cellcolor{table_col}\textbf{95.54}
        & \cellcolor{table_col}\textbf{93.68}
        & \cellcolor{table_col}\textbf{92.48}
        & \cellcolor{table_col}\textbf{97.53}
        & \cellcolor{table_col}\textbf{96.24} \\
    \bottomrule
  \end{tabular}
  }
\end{table*}

\begin{table*}[htbp]
  \centering
  \setlength{\tabcolsep}{9pt}
  \caption{Comparison of ACC (\%) under unknown distortions.}
  \label{tab:table3}
  \renewcommand{\arraystretch}{0.90}
  \resizebox{0.9\linewidth}{!}{
  \begin{tabular}{ccc|ccccccccc}
    \toprule
    \textbf{Datasets} & \textbf{Model} & \textbf{Strategy} &
    \begin{tabular}[c]{@{}c@{}}\textbf{JPEG}\\ ($Q=50$)\end{tabular} & 
    \begin{tabular}[c]{@{}c@{}}\textbf{Bright}\\ ($f=3$)\end{tabular} & 
    \begin{tabular}[c]{@{}c@{}}\textbf{Dropout}\\ ($p=0.7$)\end{tabular} & 
    \begin{tabular}[c]{@{}c@{}}\textbf{Erasing}\\ ($\sigma=0.6$)\end{tabular} & 
    \begin{tabular}[c]{@{}c@{}}\textbf{Contrast}\\ ($f=4$)\end{tabular} & 
    \begin{tabular}[c]{@{}c@{}}\textbf{Saturation}\\ ($f=3$)\end{tabular} & 
    \begin{tabular}[c]{@{}c@{}}\textbf{Cropout}\\ ($p=0.7$)\end{tabular} & 
    \begin{tabular}[c]{@{}c@{}}\textbf{Crop}\\ ($p=0.7$)\end{tabular} &
    \textbf{Avg.} \\
    \midrule

    % ==================== DIV2K ====================
    \multirow{10}{*}{\rotatebox{90}{DIV2K}}
      & \multirow{2}{*}{StegaStamp}
        & SRD & 99.89 & 77.53 & 82.75 & 78.63 & 77.05 & 61.60 & 81.17 & 84.11 & 80.34 \\
      & & \cellcolor{table_col}Meta-FC
        & \cellcolor{table_col}\textbf{99.97}
        & \cellcolor{table_col}\textbf{78.43}
        & \cellcolor{table_col}\textbf{84.34}
        & \cellcolor{table_col}\textbf{79.22}
        & \cellcolor{table_col}\textbf{79.28}
        & \cellcolor{table_col}\textbf{68.80}
        & \cellcolor{table_col}\textbf{86.39}
        & \cellcolor{table_col}\textbf{85.68}
        & \cellcolor{table_col}\textbf{82.76} \\
      \cmidrule(l){2-12}

      & \multirow{2}{*}{MBRS}
        & SRD & 60.30 & \textbf{92.70} & 99.02 & 82.45 & 96.13 & 99.71 & 91.25 & 89.53 & 88.89 \\
      & & \cellcolor{table_col}Meta-FC
        & \cellcolor{table_col}\textbf{73.65}
        & \cellcolor{table_col}92.66
        & \cellcolor{table_col}\textbf{99.74}
        & \cellcolor{table_col}\textbf{85.35}
        & \cellcolor{table_col}\textbf{96.41}
        & \cellcolor{table_col}\textbf{99.98}
        & \cellcolor{table_col}\textbf{91.54}
        & \cellcolor{table_col}\textbf{91.18}
        & \cellcolor{table_col}\textbf{91.31} \\
      \cmidrule(l){2-12}

      & \multirow{2}{*}{FIN}
        & SRD & 99.93 & 90.09 & 81.93 & 73.45 & 89.23 & 99.38 & 77.51 & 78.44 & 86.25 \\
      & & \cellcolor{table_col}Meta-FC
        & \cellcolor{table_col}\textbf{99.98}
        & \cellcolor{table_col}\textbf{90.74}
        & \cellcolor{table_col}\textbf{83.41}
        & \cellcolor{table_col}\textbf{76.06}
        & \cellcolor{table_col}\textbf{89.45}
        & \cellcolor{table_col}\textbf{99.46}
        & \cellcolor{table_col}\textbf{79.17}
        & \cellcolor{table_col}\textbf{79.57}
        & \cellcolor{table_col}\textbf{87.23} \\
      \cmidrule(l){2-12}

      & \multirow{2}{*}{SepMark}
        & SRD & 83.36 & 95.23 & 96.03 & 94.35 & 95.82 & 99.48 & 96.42 & 97.97 & 94.83 \\
      & & \cellcolor{table_col}Meta-FC
        & \cellcolor{table_col}\textbf{91.02}
        & \cellcolor{table_col}\textbf{98.05}
        & \cellcolor{table_col}\textbf{98.14}
        & \cellcolor{table_col}\textbf{98.75}
        & \cellcolor{table_col}\textbf{98.62}
        & \cellcolor{table_col}\textbf{99.93}
        & \cellcolor{table_col}\textbf{98.63}
        & \cellcolor{table_col}\textbf{99.74}
        & \cellcolor{table_col}\textbf{97.86} \\
      \cmidrule(l){2-12}

      & \multirow{2}{*}{DERO}
        & SRD & 96.32 & 92.80 & 93.86 & 89.71 & 94.09 & 99.44 & 92.64 & 94.16 & 94.13 \\
      & & \cellcolor{table_col}Meta-FC
        & \cellcolor{table_col}\textbf{98.65}
        & \cellcolor{table_col}\textbf{93.47}
        & \cellcolor{table_col}\textbf{97.64}
        & \cellcolor{table_col}\textbf{92.56}
        & \cellcolor{table_col}\textbf{96.22}
        & \cellcolor{table_col}\textbf{99.95}
        & \cellcolor{table_col}\textbf{93.59}
        & \cellcolor{table_col}\textbf{96.59}
        & \cellcolor{table_col}\textbf{96.08} \\
    \midrule

    % ==================== COCO ====================
    \multirow{10}{*}{\rotatebox{90}{COCO}}
      & \multirow{2}{*}{StegaStamp}
        & SRD & 99.85 & 73.96 & 82.32 & 79.88 & 70.70 & 59.70 & 80.85 & 84.70 & 79.00 \\
      & & \cellcolor{table_col}Meta-FC
        & \cellcolor{table_col}\textbf{99.98}
        & \cellcolor{table_col}\textbf{76.26}
        & \cellcolor{table_col}\textbf{84.41}
        & \cellcolor{table_col}\textbf{80.39}
        & \cellcolor{table_col}\textbf{74.05}
        & \cellcolor{table_col}\textbf{65.55}
        & \cellcolor{table_col}\textbf{85.63}
        & \cellcolor{table_col}\textbf{86.04}
        & \cellcolor{table_col}\textbf{81.54} \\
      \cmidrule(l){2-12}

      & \multirow{2}{*}{MBRS}
        & SRD & 60.81 & \textbf{90.78} & 99.19 & 82.94 & 96.03 & \textbf{100.00} & 91.85 & 89.72 & 88.92 \\
      & & \cellcolor{table_col}Meta-FC
        & \cellcolor{table_col}\textbf{72.44}
        & \cellcolor{table_col}90.70
        & \cellcolor{table_col}\textbf{99.77}
        & \cellcolor{table_col}\textbf{85.84}
        & \cellcolor{table_col}\textbf{96.20}
        & \cellcolor{table_col}\textbf{100.00}
        & \cellcolor{table_col}\textbf{92.37}
        & \cellcolor{table_col}\textbf{91.20}
        & \cellcolor{table_col}\textbf{91.07} \\
      \cmidrule(l){2-12}

      & \multirow{2}{*}{FIN}
        & SRD & 99.96 & 86.84 & 86.83 & 73.25 & 87.72 & 99.43 & 78.07 & 78.42 & 86.31 \\
      & & \cellcolor{table_col}Meta-FC
        & \cellcolor{table_col}\textbf{99.99}
        & \cellcolor{table_col}\textbf{88.74}
        & \cellcolor{table_col}\textbf{87.41}
        & \cellcolor{table_col}\textbf{75.36}
        & \cellcolor{table_col}\textbf{88.03}
        & \cellcolor{table_col}\textbf{99.95}
        & \cellcolor{table_col}\textbf{78.95}
        & \cellcolor{table_col}\textbf{79.60}
        & \cellcolor{table_col}\textbf{87.25} \\
      \cmidrule(l){2-12}

      & \multirow{2}{*}{SepMark}
        & SRD & 82.54 & 92.34 & 96.99 & 95.58 & 95.62 & 99.55 & 97.29 & 98.05 & 94.74 \\
      & & \cellcolor{table_col}Meta-FC
        & \cellcolor{table_col}\textbf{90.62}
        & \cellcolor{table_col}\textbf{96.55}
        & \cellcolor{table_col}\textbf{99.29}
        & \cellcolor{table_col}\textbf{99.30}
        & \cellcolor{table_col}\textbf{98.13}
        & \cellcolor{table_col}\textbf{99.99}
        & \cellcolor{table_col}\textbf{99.08}
        & \cellcolor{table_col}\textbf{99.89}
        & \cellcolor{table_col}\textbf{97.86} \\
      \cmidrule(l){2-12}

      & \multirow{2}{*}{DERO}
        & SRD & 96.28 & 91.39 & 95.20 & 90.89 & 92.02 & 99.23 & 91.78 & 94.95 & 93.97 \\
      & & \cellcolor{table_col}Meta-FC
        & \cellcolor{table_col}\textbf{98.52}
        & \cellcolor{table_col}\textbf{91.88}
        & \cellcolor{table_col}\textbf{98.25}
        & \cellcolor{table_col}\textbf{93.31}
        & \cellcolor{table_col}\textbf{94.80}
        & \cellcolor{table_col}\textbf{99.92}
        & \cellcolor{table_col}\textbf{93.45}
        & \cellcolor{table_col}\textbf{96.43}
        & \cellcolor{table_col}\textbf{95.82} \\
    \midrule

    % ==================== ImageNet ====================
    \multirow{10}{*}{\rotatebox{90}{ImageNet}}
      & \multirow{2}{*}{StegaStamp}
        & SRD & 99.88 & 75.28 & 78.45 & 79.60 & 70.82 & 65.23 & 80.17 & 85.30 & 79.34 \\
      & & \cellcolor{table_col}Meta-FC
        & \cellcolor{table_col}\textbf{99.98}
        & \cellcolor{table_col}\textbf{77.96}
        & \cellcolor{table_col}\textbf{81.90}
        & \cellcolor{table_col}\textbf{80.41}
        & \cellcolor{table_col}\textbf{74.83}
        & \cellcolor{table_col}\textbf{68.66}
        & \cellcolor{table_col}\textbf{85.47}
        & \cellcolor{table_col}\textbf{86.14}
        & \cellcolor{table_col}\textbf{81.92} \\
      \cmidrule(l){2-12}

      & \multirow{2}{*}{MBRS}
        & SRD & 65.19 & \textbf{92.57} & 99.56 & 82.61 & 96.09 & 99.77 & 92.33 & 89.92 & 89.75 \\
      & & \cellcolor{table_col}Meta-FC
        & \cellcolor{table_col}\textbf{75.19}
        & \cellcolor{table_col}92.32
        & \cellcolor{table_col}\textbf{99.82}
        & \cellcolor{table_col}\textbf{85.68}
        & \cellcolor{table_col}\textbf{96.52}
        & \cellcolor{table_col}\textbf{99.98}
        & \cellcolor{table_col}\textbf{92.96}
        & \cellcolor{table_col}\textbf{90.91}
        & \cellcolor{table_col}\textbf{91.67} \\
      \cmidrule(l){2-12}

      & \multirow{2}{*}{FIN}
        & SRD & 99.93 & 88.31 & 85.08 & 73.03 & 88.48 & 99.20 & 77.45 & 78.10 & 86.20 \\
      & & \cellcolor{table_col}Meta-FC
        & \cellcolor{table_col}\textbf{99.98}
        & \cellcolor{table_col}\textbf{89.02}
        & \cellcolor{table_col}\textbf{87.16}
        & \cellcolor{table_col}\textbf{75.83}
        & \cellcolor{table_col}\textbf{88.66}
        & \cellcolor{table_col}\textbf{99.97}
        & \cellcolor{table_col}\textbf{79.21}
        & \cellcolor{table_col}\textbf{79.28}
        & \cellcolor{table_col}\textbf{87.39} \\
      \cmidrule(l){2-12}

      & \multirow{2}{*}{SepMark}
        & SRD & 83.59 & 94.75 & 96.19 & 94.99 & 94.76 & 99.66 & 96.97 & 98.65 & 94.94 \\
      & & \cellcolor{table_col}Meta-FC
        & \cellcolor{table_col}\textbf{91.40}
        & \cellcolor{table_col}\textbf{98.20}
        & \cellcolor{table_col}\textbf{99.42}
        & \cellcolor{table_col}\textbf{99.19}
        & \cellcolor{table_col}\textbf{98.09}
        & \cellcolor{table_col}\textbf{100.00}
        & \cellcolor{table_col}\textbf{99.18}
        & \cellcolor{table_col}\textbf{99.90}
        & \cellcolor{table_col}\textbf{98.17} \\
      \cmidrule(l){2-12}

      & \multirow{2}{*}{DERO}
        & SRD & 96.45 & 94.07 & 95.10 & 90.04 & 92.25 & 99.54 & 92.62 & 94.36 & 94.30 \\
      & & \cellcolor{table_col}Meta-FC
        & \cellcolor{table_col}\textbf{98.67}
        & \cellcolor{table_col}\textbf{94.20}
        & \cellcolor{table_col}\textbf{98.55}
        & \cellcolor{table_col}\textbf{92.78}
        & \cellcolor{table_col}\textbf{94.48}
        & \cellcolor{table_col}\textbf{99.95}
        & \cellcolor{table_col}\textbf{93.72}
        & \cellcolor{table_col}\textbf{96.41}
        & \cellcolor{table_col}\textbf{96.10} \\
    \bottomrule
  \end{tabular}
  }
\end{table*}

\subsection{Robustness against Known Distortions}
In this section, we compare the performance of five watermarking models trained using the SRD and Meta-FC under known distortions. All models are trained with Identity and a set of 8 common distortions: Gaussian Blur (GB) ($\sigma=2$), Salt-and-Pepper Noise (SPN) ($\sigma=0.04$), Crop ($p=0.2$), JPEG ($\mathcal{Q}=50$), Median Blur (MB) ($w=5$), Gaussian Noise (GN) ($\sigma=0.04$), Brightness ($f\in[0.85, 1.15]$), and Dropout ($p=0.3$). To ensure a fair comparison, we adjust the embedding strength for each model so that visual quality remains consistent with the two training strategies. Figure \ref{fig:figure3} illustrates the visual results for each model, including the watermarked images, their corresponding residuals, and the distorted image.

\subsubsection{High-intensity Distortions.} We first evaluate the robustness of models under Identity and 8 high-intensity distortions, including GB ($\sigma=6$), SPN ($\sigma=0.15$), Crop ($p=0.7$), JPEG ($\mathcal{Q}=30$), MB ($w=7$), GN ($\sigma=0.08$), Brightness ($f=4$), and Dropout ($p=0.7$). Table \ref{tab:table1} shows the specific results of robustness. 
In most cases, models trained with Meta-FC outperform those trained with SRD. Notably, Meta-FC aims for broad generalization, while SRD tends to overfit certain attacks. Consequently, under certain attacks, Meta-FC shows slightly lower performance, but the difference remains within 1\% ACC. On average, Meta-FC achieves a 1.57\% improvement over SRD, demonstrating the robustness of our method against high-intensity distortions.

\subsubsection{Combined Distortions.} In practical scenarios, watermarked images may be subject to multiple types of distortions during transmission. Therefore, it is also essential to evaluate the robustness of the watermark against combined distortions. To test the robustness of Meta-FC and SRD, we simulate a set of combined distortions using MB ($w=5$), GN ($\sigma=0.05$), SPN ($\sigma=0.08$), Crop ($p=0.3$), and JPEG ($\mathcal{Q}=50$). The specific combined distortions include: MB \& JPEG, MB \& SPN, MB \& GN, GN \& SPN, GN \& JPEG, MB \& SPN \& GN, and MB \& JPEG \& Crop.

As reported in Table \ref{tab:table2}, Meta-FC consistently demonstrates superior robustness compared to SRD under combined distortions, except for the FIN under the MB\&GN distortion.
On average, it achieves ACC improvements of 0.63\%, 7.92\%, 1.62\%, 12.79\%, and 1.23\% over SRD on StegaStamp, MBRS, FIN, SepMark, and DERO, respectively.
These results highlight the advantage of introducing meta-learning with feature consistency loss into the training process. 
As meta-learning promotes the discovery of parameters that remain effective under diverse distortions, the feature consistency loss further regularizes the model to produce stable decoding features across different distortions. This joint optimization enables the model to form distortion-consistent representations, thereby significantly enhancing its robustness against combined distortions.

\subsection{Robustness against Unknown Distortions} 
Although the noise pool covers a wide range of common distortions, watermarked images may be subject to previously unknown distortions (not included in the noise pool). Therefore, a reliable watermarking model should generalize well to unknown distortions. 
To evaluate the generalization of Meta-FC and SRD, we train each model with Identity and a set of four training distortions: GB ($\sigma=2$), SPN ($\sigma=0.04$), GN ($\sigma=0.04$), and MB ($w=5$). The trained models are then evaluated on eight unknown distortions from different classes, including JPEG ($\mathcal{Q}=50$), Brightness ($f=3$), Dropout ($p=0.7$), Erasing ($\sigma=0.6$), Contrast ($f=4$), Saturation($f=3$), Cropout($p=0.7$), and Crop($p=0.7$). 
To ensure a fair comparison, the embedding strength of the models is adjusted to maintain consistent image quality under Meta-FC and SRD. The resulting PSNR (dB) for StegaStamp, MBRS, FIN, SepMark, and DERO are set to 31, 32, 33, 32, and 34.

Table \ref{tab:table3} reports the ACC results under unknown distortions. Meta-FC consistently outperforms SRD, achieving average ACC improvements of 2.51\%, 2.16\%, 1.04\%, 3.12\%, and 1.87\% on StegaStamp, MBRS, FIN, SepMark, and DERO, respectively. 
The superior generalization performance of Meta-FC against unknown distortions is attributed to the meta-testing phase, which evaluates the model using ``unknown" distortions.
This mechanism effectively simulates real-world scenarios involving unknown distortions and encourages the model to learn generalized, distortion-invariant representations that are transferable beyond the training noise pool.

\subsection{Ablation Study} 
In this section, we conduct ablation studies on the key components of the proposed Meta-FC method, as well as on the size of the noise pool, to better understand their contributions to overall performance improvements. All ablation experiments are performed on the test set, where various types of noise are randomly applied to compute the ACC while keeping the PSNR of watermarked images consistent. 

\begin{table}[t]
  \centering
  \caption{Ablation study on different parts in our proposed Meta-FC architecture.}
  % \scalebox{0.9}{
  \resizebox{\linewidth}{!}{
  \begin{tabular}{c|ccc}
    \toprule
    \multirow{2}[2]{*}{\textbf{Strategy}} & \multicolumn{3}{c}{\textbf{ACC(\%) $\uparrow$}} \\ 
    \cmidrule(lr){2-4}
    & \textbf{High-intensity} & \textbf{Combined} & \textbf{Unknown} \\
    \midrule
    Mate-FC \textit{w/o} meta-train \& FC & 94.34 ($\downarrow$ 1.59) & 87.94 ($\downarrow$ 4.71) & 88.55 ($\downarrow$ 2.38) \\
    Mate-FC \textit{w/o} meta-test \& FC  & 95.27 ($\downarrow$ 0.66) & 90.58 ($\downarrow$ 2.07) & 88.96 ($\downarrow$ 1.97) \\
    Mate-FC \textit{w/o} FC               & 95.56 ($\downarrow$ 0.37) & 90.44 ($\downarrow$ 2.21) & \textbf{90.99} ($\uparrow$ 0.06) \\
    \rowcolor{table_col}
    Meta-FC                      & \textbf{95.93} & \textbf{92.65} & 90.93 \\
    \bottomrule
  \end{tabular}
  }
  \vspace{-5pt}
  \label{tab:table4}
\end{table}

We conduct ablation experiments on the proposed Meta-FC method to investigate the contributions of its three core components: meta-train, meta-test, and feature consistency loss. 
As shown in Table \ref{tab:table4}, \textbf{Mate-FC \textit{w/o} meta-train \& FC} refers to the variant that includes only the meta-testing phase, \textbf{Mate-FC \textit{w/o} meta-test \& FC} includes only the meta-training phase, and \textbf{Mate-FC \textit{w/o} FC} refers to the full meta-learning process without the feature consistency loss. Note that \textbf{Mate-FC \textit{w/o} meta-train \& FC} variant is effectively identical to the widely used SRD baseline.
The results indicate that the meta-training phase plays a more critical role than the meta-testing phase. Furthermore, the complete meta-learning architecture achieves a higher ACC than either individual phase alone. Introducing the feature consistency loss, Meta-FC yields improved generalization performance on combined distortions, outperforming the Meta-learning variant by approximately 2.21\%.

\subsection{Time Complexity Analysis}
In Table \ref{tab:table5}, we report the training time and corresponding ACC (average of high-intensity, combined, and unknown distortions) of Meta-FC and SRD. All models are trained on an NVIDIA RTX 3090 GPU. The results show that Meta-FC requires approximately $0.6$ times additional training time compared with SRD. Considering that Meta-FC achieves a higher ACC, this additional cost is well justified.

\begin{table}[t]
\centering
\caption{Comparison of training time for SRD and Meta-FC.}
\renewcommand{\arraystretch}{1}
\resizebox{\linewidth}{!}{
\begin{tabular}{@{}c|cl|cl|cl|cl|cl@{}}
\toprule
\multirow{2}{*}{\textbf{Strategy}} & \multicolumn{2}{c|}{\textbf{StegaStamp}} & \multicolumn{2}{c|}{\textbf{MBRS}}      & \multicolumn{2}{c|}{\textbf{FIN}}       & \multicolumn{2}{c|}{\textbf{SepMark}}   & \multicolumn{2}{c}{\textbf{DERO}}       \\ \cmidrule(l){2-11} 
                                   & \textbf{Time (h)}            & \textbf{ACC (\%)}                & \textbf{Time (h)}       & \textbf{ACC (\%)}           & \textbf{Time (h)}       & \textbf{ACC (\%)}           & \textbf{Time (h)}       & \textbf{ACC (\%)}           & \textbf{Time (h)}       & \textbf{ACC (\%)}           \\ \midrule
SRD                                & \textbf{1.81}      & 94.67               & \textbf{2.07} & 85.93          & \textbf{1.68} & 90.80          & \textbf{1.99} & 89.19          & \textbf{1.77} & 95.05          \\
\rowcolor{table_col}
Meta-FC                            & 2.36               & \textbf{96.87}      & 3.37          & \textbf{90.40} & 2.86          & \textbf{92.35} & 3.12          & \textbf{95.48} & 2.53          & \textbf{96.59} \\ \bottomrule
\end{tabular}
}
\label{tab:table5}
\end{table}

\section{Conclusion}
In this paper, we propose a novel training strategy, Meta-FC, to overcome the inherent limitations of the SRD paradigm in deep learning-based watermarking.
The proposed Meta-FC is plug-and-play and can be seamlessly integrated with any END-based watermarking model. By simulating alternating training with known and ``unknown'' distortions, Meta-FC encourages the model to discover parameter configurations with stable activations across diverse distortions.  
Furthermore, a feature consistency loss is introduced to encourage the decoder to produce similar representations for the same image under different distortions, improving its robustness.
Extensive experiments demonstrate that Meta-FC consistently outperforms SRD in both robustness and generalization across all evaluated distortion scenarios.
Nevertheless, it is important to note that a model’s architecture fundamentally limits how far its generalization can extend. Meta-FC significantly improves robustness within these architectural constraints, but full resistance to completely unseen attack types is beyond the capability of any training strategy alone. Thus, future work may focus on designing stronger model architectures, which can further amplify the benefits provided by Meta-FC.